\documentclass{article}
\usepackage{times}
\usepackage{epsfig}
\usepackage{graphicx}
\usepackage{subcaption} 
\usepackage{amsmath}
\usepackage{amssymb}
\usepackage{algorithm}
\usepackage{adjustbox}
\usepackage{diagbox}
\usepackage[noend]{algpseudocode}
\usepackage[utf8]{inputenc}

\title{Quantizing deep convolutional networks for efficient inference: A whitepaper}
\author{Raghuraman Krishnamoorthi \\
raghuramank@google.com}
\date{June 2018}
\begin{document}
\maketitle
\tableofcontents
\newpage
\begin{abstract}

 We present an overview of techniques for quantizing convolutional neural networks for inference with integer weights and activations. 

 \begin{enumerate}
     \item Per-channel quantization of weights and per-layer quantization of activations to 8-bits of precision post-training produces classification accuracies within 2\% of floating point networks for a wide variety of CNN architectures (section \ref{sec:PostTrnQtz}).
     
     \item Model sizes can be reduced by a factor of 4 by quantizing weights to 8-bits, even when 8-bit arithmetic is not supported. This can be achieved with simple, post training quantization of weights (section \ref{sec:PostTrnQtz}).
     
     \item We benchmark latencies of quantized networks on CPUs and DSPs and observe a speedup of 2x-3x for quantized implementations compared to floating point on CPUs. Speedups of up to 10x are observed on specialized processors with fixed point SIMD capabilities, like the Qualcomm QDSPs with HVX (section \ref{sec:RTM}).
     
     \item Quantization-aware training can provide further improvements, reducing the gap to floating point to 1\% at 8-bit precision. Quantization-aware training also allows for reducing the precision of weights to four bits with accuracy losses ranging from 2\% to 10\%, with higher accuracy drop for smaller networks (section \ref{sec:QtzAwareTrn}).
     
     \item We introduce tools in TensorFlow and TensorFlowLite for quantizing convolutional networks (Section \ref{sec:QtzTrn}).
    
\item  We review best practices for quantization-aware training to obtain high accuracy with quantized weights and activations (section \ref{TrnPractices}).
 
\item  We recommend that per-channel quantization of weights and per-layer quantization of activations be the preferred quantization scheme for hardware acceleration and kernel optimization. We also propose that future processors and hardware accelerators for optimized inference support precisions of 4, 8 and 16 bits (section \ref{sec:NNRec}).
 
\end{enumerate}

\end{abstract}

\section{Introduction} \label{Intro}
Deep networks are increasingly used for applications at the edge. Devices at the edge typically have lower compute capabilities and are constrained in memory and power consumption. It is also necessary to reduce the amount of communication to the cloud for transferring models to the device to save on power and reduce network connectivity requirements. Therefore, there is a pressing need for techniques to optimize models for reduced model size, faster inference and lower power consumption.

There is extensive research on this topic with several approaches being considered: One approach is to build efficient models from the ground up \cite{MobilenetV2},\cite{AHoward17} and \cite{IandolaMAHDK16}. Another technique is to reduce the model size by applying quantization, pruning and compression techniques \cite{BJacob17}, \cite{CourbariauxBD15} and \cite{HanMD15}. Faster inference has been achieved by having efficient kernels for computation in reduced precision like GEMMLOWP \cite{GEMMLOWP}, Intel MKL-DNN \cite{IntcMKL} , ARM CMSIS \cite{ARMCMSIS}, Qualcomm SNPE \cite{QcomSNPE}, Nvidia TensorRT \cite{TensorRT} and custom hardware for fast inference \cite{SzeCYE17}, \cite{NvidiaDLA} and \cite{HanEiE}.

One of the simpler ways to reduce complexity of any model is to reduce the precision requirements for the weights and activations. This approach has many advantages:
\begin{itemize}
\item It is broadly applicable across a range of models and use cases. One does not need to develop a new model architecture for improved speed.
In many cases, one can start with an existing floating point model and quickly quantize it to obtain a fixed point quantized model with almost no accuracy loss, without needing to re-train the model.
Multiple hardware platforms and libraries support fast inference with quantized weights and activations, so there is no need to wait for new hardware development.
\item Smaller Model footprint: With 8-bit quantization, one can reduce the model size a factor of 4, with negligible accuracy loss. This can be done without needing any data as only the weights are quantized. This also leads to faster download times for model updates. 
\item Less working memory and cache for activations: Intermediate computations are typically stored in cache for reuse by later layers of a deep network and reducing the precision at which this data is stored leads to less working memory needed. Having lower precision weights and activations allows for better cache reuse.
\item Faster computation: Most processors allow for faster processing of 8-bit data. 
\item Lower Power: Moving 8-bit data is 4 times more efficient than moving 32-bit floating point data. In many deep architectures, memory access can dominate power consumption [2]. Therefore reduction in amount of data movement can have a significant impact on the power consumption.
\end{itemize}
All the factors above translate into faster inference, with a typical speedup of 2-3x due to the reduced precision for both memory accesses and computations. Further improvements in speed and power consumption are possible with processors and hardware accelerators optimized for low precision vector arithmetic.

\section{Quantizer Design}
 In this section, we review different design choices for uniform quantization.
\subsection{Uniform Affine Quantizer}
Consider a floating point variable with range $(x_{min},x_{max})$ that needs to be quantized to the range $(0,N_{levels}-1)$ where $N_{levels}=256$ for 8-bits of precision. We derive two parameters: Scale ($\Delta $) and Zero-point($z$) which map the floating point values to integers (See \cite{NNAPI}). The scale specifies the step size of the quantizer and floating point zero maps to zero-point \cite{BJacob17}. Zero-point is an integer, ensuring that zero is quantized with no error. This is important to ensure that common operations like zero padding do not cause quantization error.

For one sided distributions, therefore, the range $(x_{min},x_{max})$ is relaxed to include zero. For example, a floating point variable with the range (2.1,3.5) will be relaxed to the range (0,3.5) and then quantized. Note that this can cause a loss of precision in the case of extreme one-sided distributions.

Once the scale and zero-point are defined, quantization proceeds as follows:
\begin{align}
    x_{int}&= round\big(\frac{x}{\Delta}\big)+z \\
    x_Q&=clamp(0,N_{levels} - 1, x_{int}) 
\end{align}
where \begin{align*}
    clamp(a,b,x)&=a &x\leq a \\
    &=x &a \leq x \leq b \\
    &=b &x \geq b
\end{align*}
The de-quantization operation is:
\begin{align} 
 x_{float}=(x_Q  - z)\Delta \label{eq:dequant}
\end{align}

While the uniform affine quantizer allows for storing weights and activations at 8-bits of precision, there is an additional cost due to the zero-point.
Consider a 2D convolution between a weight and an activation:

\begin{align}
    y(k,l,n)&=\Delta_w \Delta_x conv(w_Q(k,l,m;n)-z_{w}, x_Q(k,l,m)-z_{x}) \\
    y(k,l,n)&=conv(w_Q(k,l,m;n),x_Q(k,l,m))-z_w\sum_{k=0}^{K-1} \sum_{l=0}^{K-1} \sum_{m=0}^{N-1} x_Q(k,l,m) \\
    &-z_x\sum_{k=0}^{K-1} \sum_{l=0}^{K-1} \sum_{m=0}^{N-1} w_Q(k,l,m;n) +z_x z_w
\end{align}

A naive implementation of convolution, by performing the addition of zero-point prior to the convolution, leads to a 2x to 4x reduction in the throughput due to wider (16/32-bit) operands. One can do better by using the equation above and noting that the last term is a constant and each of the other terms requires N multiplies, which is 3x more operations than the 8-bit dot product. This can be further improved by noting that the weights are constant at inference and by noting that the sum over activations is identical for all convolutional kernels of the same size. However, this requires optimizing convolution kernels. For an indepth discussion, please see \cite{GEMMLOWPQtz}.
 
\subsection{Uniform symmetric quantizer}
A simplified version of the affine quantizer is the symmetric quantizer, which restricts zero-point to 0. With the symmetric quantizer, the conversion operations simplify to:
\begin{align}
    x_{int}&= round\big(\frac{x}{\Delta}\big) & \\
    x_Q&=clamp(-N_{levels}/2,N_{levels}/2 - 1, x_{int})& \text{if signed}  \\
    x_Q&=clamp(0,N_{levels} - 1, x_{int}) &\text{if un-signed} 
\end{align}

For faster SIMD implementation, we further restrict the ranges of the weights. In this case, the clamping is modified to:
\begin{align}
    x_Q&=clamp(-(N_{levels}/2 - 1),N_{levels}/2 - 1, x_{int})& \text{if signed}  \\
    x_Q&=clamp(0,N_{levels} - 2, x_{int}) &\text{if un-signed} 
\end{align}
 Please see \cite{BJacob17}, Appendix B for more details.

The de-quantization operation is:
\begin{equation*}
 x_{out}=x_Q\Delta
\end{equation*}
\subsection{Stochastic quantizer}
 Stochastic quantization models the quantizer as an additive noise, followed by rounding. The stochastic quantizer is given by:
\begin{align*}
    x_{int}& = round\big(\frac{x+\epsilon}{\Delta}\big)+z, \qquad \epsilon \sim Unif(-\frac{1}{2}, \frac{1}{2}) \\
     x_Q&=clamp(0,N_{levels} - 1, x_{int}) 
\end{align*}

The de-quantization operation is given by equation \ref{eq:dequant}.
Note that in expectation, the stochastic quantizer reduces to a pass-through of the floating point weights, with saturation for values outside the range. Therefore, this function is well behaved for purposes of calculating gradients. We do not consider stochastic quantization for inference as most inference hardware does not support it. 

\subsection{Modeling simulated quantization in the backward pass} \label{sec:modelQuant}
\begin{figure}
[!htbp]
\begin{center}
    \includegraphics[width=0.8\linewidth]{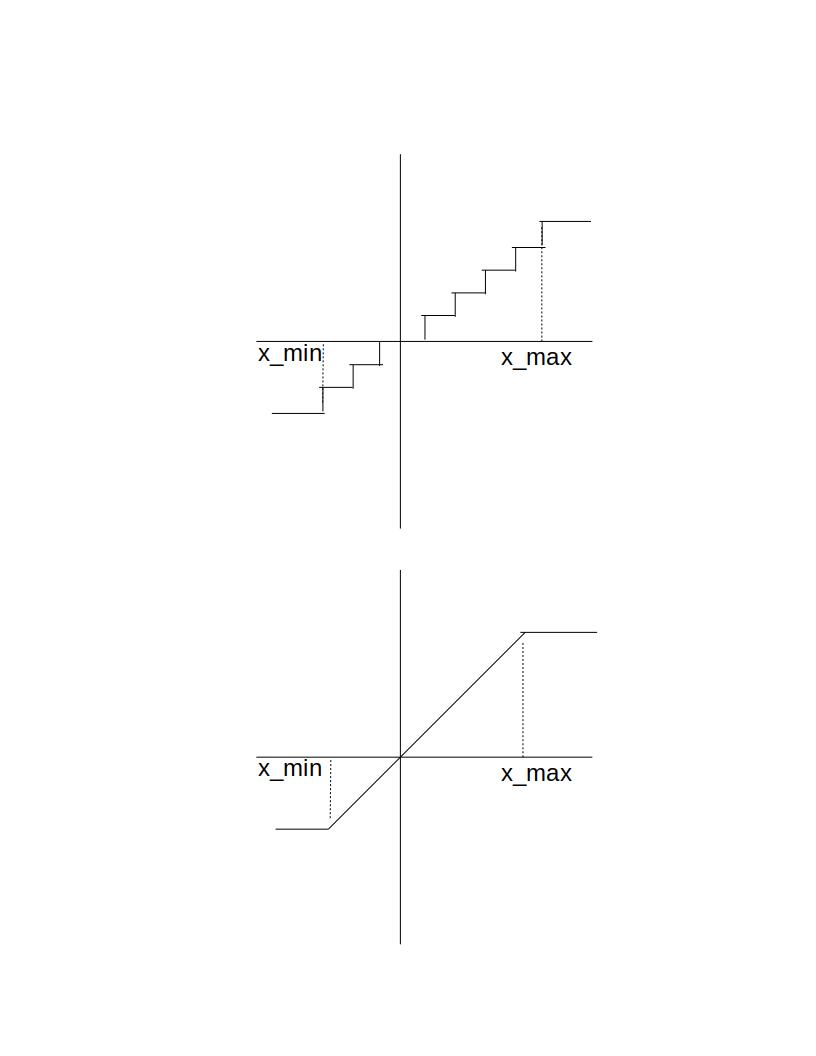}
\end{center}
   \caption{Simulated Quantizer (top), showing the quantization of output values. Approximation for purposes of derivative calculation (bottom).}
\label{fig:SimQuant}
\end{figure}

For Quantization-aware training, we model the effect of quantization using simulated quantization operations, which consist of a quantizer followed by a de-quantizer, i.e, 
\begin{align} 
x_{out}&=SimQuant(x) \\
       &=\Delta\: clamp\big(0,N_{levels}-1,round(\frac{x}{\Delta})-z\big)
\end{align}
Since the derivative of a simulated uniform quantizer function is zero almost everywhere, approximations are required to model a quantizer in the backward pass. An approximation that has worked well in practice (see \cite{CourbariauxBD15}) is to model the quantizer as specified in equation \ref{eq:ste} for purposes of defining its derivative (See figure \ref{fig:SimQuant}).

\begin{equation}
    x_{out}=clamp(x_{min},x_{max},x) \label{eq:ste}
\end{equation}
 
The backward pass is modeled as a "straight through estimator" (see \cite{CourbariauxBD15}). Specifically,
\begin{equation}
\delta_{out}=\delta_{in}I_{x \in S}  S:{x: x_{min} \leq x \leq x_{max}}
\end{equation}

where $\delta_{in} = \frac{\partial L}{\partial w_{out}} $ is the backpropagation error of the loss with respect to the simulated quantizer output.

\subsection{Determining Quantizer parameters}
The quantizer parameters can be determined using several criteria. For example, TensorRT \cite{TensorRT} minimizes the KL divergence between the original and quantized distributions to determine the step size. In this work, we adopt simpler methods. For weights, we use the actual minimum and maximum values to determine the quantizer parameters. For activations, we use the moving average of the minimum and maximum values across batches to determine the quantizer parameters. For post training quantization approaches, one can improve the accuracy of quantized models by careful selection of quantizer parameters.

\subsection{Granularity of quantization}
 We can specify a single quantizer (defined by the scale and zero-point) for an entire tensor, referred to as per-layer quantization.  Improved accuracy can be obtained by adapting the quantizer parameters to each kernel within the tensor \cite{APol18}. For example, the weight tensor is 4 dimensional and is a collection of 3 dimensional convolutional kernels, each responsible for producing one output feature map. per-channel quantization has a different scale and offset for each convolutional kernel. We do not consider per-channel quantization for activations as this would complicate the inner product computations at the core of conv and matmul operations. Both per-layer and per-channel quantization allow for efficient dot product and convolution implementation as the quantizer parameters are fixed per kernel in both cases.
 
\section {Quantized Inference: Performance and Accuracy}
\label{sec:QtzTrn}
 Quantizing a model can provide multiple benefits as discussed in section \ref{Intro}. We discuss multiple approaches for model quantization and show the performance impact for each of these approaches.
 \begin{figure*}
[!htbp]
\begin{center}
    \includegraphics[width=1.0\linewidth]{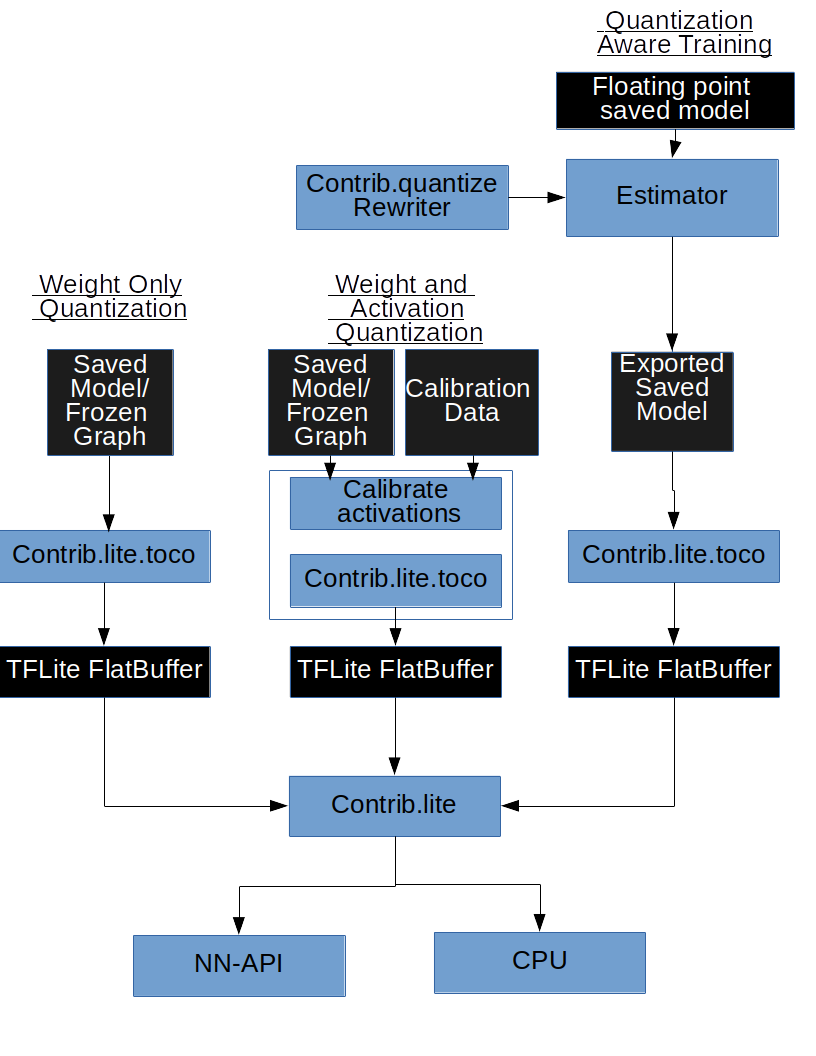}
\end{center}
   \caption{Overview of schemes for model quantization: One can quantize weights post training (left) or quantize weights and activations post training (middle). It is also possible to perform quantization aware training for improved accuracy}
\label{fig:HighLevelDiagram}
\end{figure*}

\subsection{Post Training Quantization}
\label{sec:PostTrnQtz}
In many cases, it is desirable to reduce the model size by compressing weights and/or quantize both weights and activations for faster inference, without requiring to re-train the model. Post Training quantization techniques are simpler to use and allow for quantization with limited data. In this section, we study different quantization schemes for weight only quantization and for quantization of both weights and activations. We show that per-channel quantization with asymmetric ranges produces accuracies close to floating point across a wide range of networks.

\subsubsection{Weight only quantization}
A simple approach is to only reduce the precision of the weights of the network to 8-bits from float. Since only the weights are quantized, this can be done without requiring any validation data (See figure \ref{fig:HighLevelDiagram}). A simple command line tool can convert the weights from float to 8-bit precision. This setup is useful if one only wants to reduce the model size for transmission and storage and does not mind the cost of performing inference in floating point.

\subsubsection{Quantizing weights and activations}
 One can quantize a floating point model to 8-bit precision by calculating the quantizer parameters for all the quantities to be quantized. Since activations need to be quantized, one needs calibration data and needs to calculate the dynamic ranges of activations. (See figure \ref{fig:HighLevelDiagram}) Typically, about 100 mini-batches are sufficient for the estimates of the ranges of the activation to converge. 

\subsubsection{Experiments}
For evaluating the tradeoffs with different quantization schemes, we study the following popular networks and evaluate the top-1 classification accuracy. Table \ref{Table:NetworkSize} shows the wide variation in model size and accuracy across these networks. We note that Mobilenet-v1 \cite{AHoward17} and Mobilenet-v2\cite{MobilenetV2}  architectures use separable depthwise and pointwise convolutions with Mobilenet-v2 also using skip connections.
Inception-v3 \cite{Szegedy15} and NasNet \cite{NasNet} use network in network building blocks with NasNet determining the architecture via reinforcement learning techniques.
Resnets \cite{ResnetV1He} pioneered the idea of skip connections and consist of multiple blocks each making residual corrections to the main path with no transformations. Resnet-v2 \cite{Resnetv2He} is an enhancement to the resnet architecture using pre-activation layers for improved accuracy. Note that all results are obtained using simulated quantization of weights and activations.

\begin{table}[!htbp]
\begin{center}
  \begin{tabular}{|l|p{2cm}|p{2cm}|}
  \hline

Network &Model Parameters & Top-1 Accuracy on ImageNet (fp32) \\ \hline
Mobilenet\_V1\_0.25\_128 &0.47M & 0.415  \\ \hline
Mobilenet\_V2\_1\_224 & 3.54M & 0.719 \\ \hline
Mobilenet\_V1\_1\_224 & 4.25M & 0.709 \\ \hline
Nasnet\_Mobile & 5.3M & 0.74 \\ \hline
Mobilenet\_V2\_1.4\_224 & 6.06M & 0.749 \\ \hline
Inception\_V3 &23.9M & 0.78 \\ \hline
Resnet\_v1\_50  & 25.6M & 0.752 \\ \hline
Resnet\_v2\_50  & 25.6M & 0.756 \\ \hline
Resnet\_v1\_152 & 60.4M & 0.768 \\ \hline
Resnet\_v2\_152 & 60.4M & 0.778 \\ \hline
  \end{tabular}
\end{center}
\caption{Deep Convolutional networks: Model size and accuracy}
\label{Table:NetworkSize}    
\end{table}

\paragraph{Weight only quantization:} 
We first quantize only the weights post training and leave the activations un-quantized. From figure \ref{Table:WtOnlyQuant}, we note that per-channel quantization is required to ensure that the accuracy drop due to quantization is small, with asymmetric, per-layer quantization providing the best accuracy.
\begin{table}[!htbp]
\begin{center}
  \begin{tabular}{|l|p{2cm}|p{2cm}|p{2cm}|p{2cm}|}
  \hline
  
Network & Asymmetric, per-layer & Symmetric , per-channel & Asymmetric, per-channel & Floating Point \\ \hline
Mobilenet\-v1\_1\_224 &0.001 &0.591 &0.704& 0.709 \\ \hline
Mobilenet\-v2\_1\_224 &0.001 &0.698 &0.698& 0.719 \\ \hline
Nasnet\-Mobile& 0.722&0.721 & 0.74 &0.74 \\ \hline
Mobilenet\-v2\_1.4\_224 &0.004 &0.74 &0.74 & 0.749 \\ \hline
Inception\-v3 & 0.78& 0.78 & 0.78 & 0.78 \\ \hline
Resnet\_v1\_50 & 0.75 & 0.751 &0.752 & 0.752 \\ \hline
Resnet\_v2\_50 & 0.75 & 0.75 &0.75 & 0.756 \\ \hline
Resnet\_v1\_152 & 0.766 & 0.763 &0.762 & 0.768 \\ \hline
Resnet\_v2\_152 & 0.761 & 0.76 &0.77 & 0.778 \\ \hline

 \end{tabular}
\end{center}
\caption{Weight only quantization: per-channel quantization provides good accuracy, with asymmetric quantization providing close to floating point accuracy.}
\label{Table:WtOnlyQuant}    
\end{table}

\paragraph{Weight and Activation Quantization:} 

Next, we quantize weights and activations to 8-bits, with per-layer quantization for activations. For weights we consider both symmetric and asymmetric quantizers at granularities of both a layer and a channel. We first show results for Mobilenet\-v1 networks and then tabulate results across a broader range of networks.

\begin{figure}
[!htbp]
\begin{center}
    \includegraphics[width=1.0\linewidth]{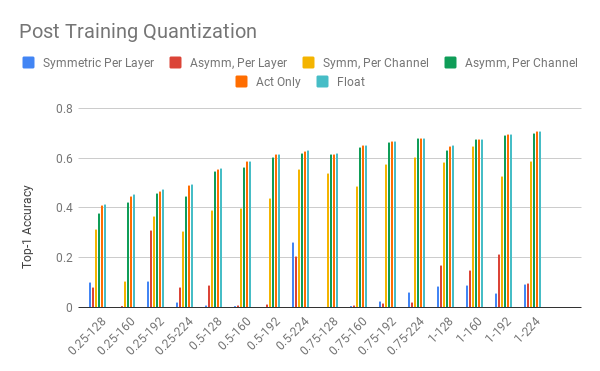}
\end{center}
   \caption{Comparison of post training weight and activation quantization schemes:Mobilenet-v1}
\label{fig:PostTrnQtzMobilenetV1}
\end{figure}

\begin{figure}
[!htbp]
\begin{center}
    \includegraphics[width=12cm,height=6cm]{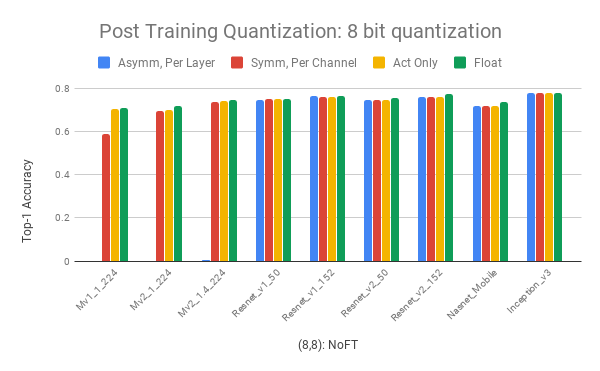}
\end{center}
   \caption{Comparison of post training quantization schemes}
\label{fig:PostTrnQtzSurvey}
\end{figure}

We also compare the post training quantization accuracies of popular convolutional networks: Inception-V3, Mobilenet-V2, Resnet-v1-50, Resnet-v1-152, Resnet-v2-50, Resnet-v2-152 and Nasnet-mobile on ImageNet in figure \ref{fig:PostTrnQtzSurvey}.
\begin{table}[!htbp]
\begin{center}
  \begin{tabular}{|l|p{2cm}|p{2cm}|p{2cm}|p{2cm}|p{2cm}|}
  \hline
  
Network & Asymmetric, per-layer & Symmetric , per-channel & Asymmetric, per-channel &  Activation Only & Floating Point \\ \hline
Mobilenet-v1\_1\_224 &0.001 &0.591 &0.703&0.708& 0.709 \\ \hline
Mobilenet-v2\_1\_224 &0.001 &0.698 &0.697&0.7& 0.719 \\ \hline
Nasnet-Mobile& 0.722&0.721 & 0.74 &0.74&0.74 \\ \hline
Mobilenet-v2\_1.4\_224 &0.004 &0.74 &0.74 &0.742& 0.749 \\ \hline
Inception-v3 & 0.78& 0.78 & 0.78 & 0.78&0.78 \\ \hline
Resnet-v1\_50 & 0.75 & 0.751 &0.751 &0.751& 0.752 \\ \hline
Resnet-v2\_50 & 0.75 & 0.75 &0.75 & 0.75&0.756 \\ \hline
Resnet-v1\_152 & 0.766 & 0.762 &0.767 &0.761& 0.768 \\ \hline
Resnet-v2\_152 & 0.761 & 0.76 &0.76 &0.76& 0.778 \\ \hline

 \end{tabular}
\end{center}
\caption{Post training quantization of weights and activations: per-channel quantization of weights and per-layer quantization of activations works well for all the networks considered, with asymmetric quantization providing slightly better accuracies.}
\label{Table:WtActQuant}    
\end{table}

We make the following observations:
\begin{enumerate}
\item Per-channel quantization can provide good accuracy and can be a good baseline for post training quantization of weights and activations, with asymmetric quantization providing close to floating point accuracy for all networks.

\item Activations can be quantized to 8-bits with almost no loss in accuracy. The dynamic ranges of the activations are low due to a combination of:
\begin{enumerate}
\item Batch normalization with no scaling: Used in Inception-V3, which ensures that the activations of all feature maps have zero mean and unit variance.
\item ReLU6: Used in Mobilenet-V1, which restricts the activations to be in a fixed range (0,6) for all feature maps, thereby removing large dynamic range variations.
\end{enumerate}
\item Networks with more parameters like Resnets and Inception-v3 are more robust to quantization compared to Mobilenets which have fewer parameters.
\item There is a large drop when weights are quantized at the granularity of a layer, particularly for Mobilenet architectures.
\item Almost all the accuracy loss due to quantization is due to weight quantization. 
\end{enumerate}

Weight quantization at the granularity of a layer causes large accuracy drops primarily due to batch normalization, which causes extreme variation in dynamic range across convolution kernels in a single layer. Appendix A has more details on the impact of batch normalization. Per-channel quantization side-steps this problem by quantizing at the granularity of a kernel, which makes the accuracy of per-channel quantization independent of the batch-norm scaling. However, the activations are still quantized with per-layer symmetric quantization. 

Note that other approaches like weight regularization can also improve the accuracy of quantization post training, please see \cite{QcomMobilenet}.

\subsection{Quantization Aware Training}
 \label{sec:QtzAwareTrn}

Quantization aware training models quantization during training and can provide higher accuracies than post quantization training schemes. In this section, we describe how quantization is modeled during training and describe how this can be easily done using automatic quantization tools in TensorFlow. We also evaluate the accuracies obtained for different quantization schemes with quantization aware training and show that even per-layer quantization schemes show high accuracies post training at 8-bits of precision. We also show that at 4 bit precision, quantization aware training provides significant improvements over post training quantization schemes.

We model the effect of quantization using simulated quantization operations on both weights and activations.
For the backward pass, we use the straight through estimator (see section \ref{sec:modelQuant}) to model quantization. Note that we use simulated quantized weights and activations for both forward and backward pass calculations. However, we maintain weights in floating point and update them with the gradient updates. This ensures that minor gradient updates gradually update the weights instead of underflowing. The updated weights are quantized and used for subsequent forward and backward pass computation. For SGD, the updates are given by:

\begin{align}
w_{float}& =w_{float}-\eta \frac{\partial L}{\partial w_{out}}.I_{w_{out} \in (w_{min},w_{max})} \\
w_{out}& =SimQuant(w_{float})
\end{align}

Quantization aware training is achieved by automatically inserting simulated quantization operations in the graph at both training and inference times using the quantization library at \cite{TFQuantize} for Tensorflow \cite{tensorflow2015-whitepaper}. We follow the approach outlined in \cite{BJacob17} closely, with additional enhancements on handling batch normalization and in modeling quantization in the backward pass. A simple one-line change to the training or evaluation code automatically inserts simulated quantization operations into the training or eval graph.
 
 For training, the code snippet is:
\begin{verbatim}
# Build forward pass of model.
...
loss = tf.losses.get_total_loss()

# Call the training rewrite which rewrites the graph in-place
# with FakeQuantization nodes and folds batchnorm for training.
# One can either fine tune an existing floating point model
# or train from scratch. quant_delay controls the onset 
# of quantized training.

tf.contrib.quantize.create_training_graph(quant_delay=2000000)

# Call backward pass optimizer as usual.
optimizer = tf.train.GradientDescentOptimizer(learning_rate)
optimizer.minimize(loss)

\end{verbatim}

For evaluation, the code snippet is given below:
\begin{verbatim}
# Build eval model
…
logits, end_points = network_model(inputs,...)

# Call the eval rewrite which rewrites the graph in-place
# with FakeQuantization nodes and fold batchnorm for eval.
tf.contrib.quantize.create_eval_graph()



\end{verbatim}

 The high level conversion process is shown in figure \ref{fig:HighLevelDiagram}.

\begin{figure}
\begin{subfigure}[h]{0.8\linewidth}
\includegraphics[width=\linewidth]{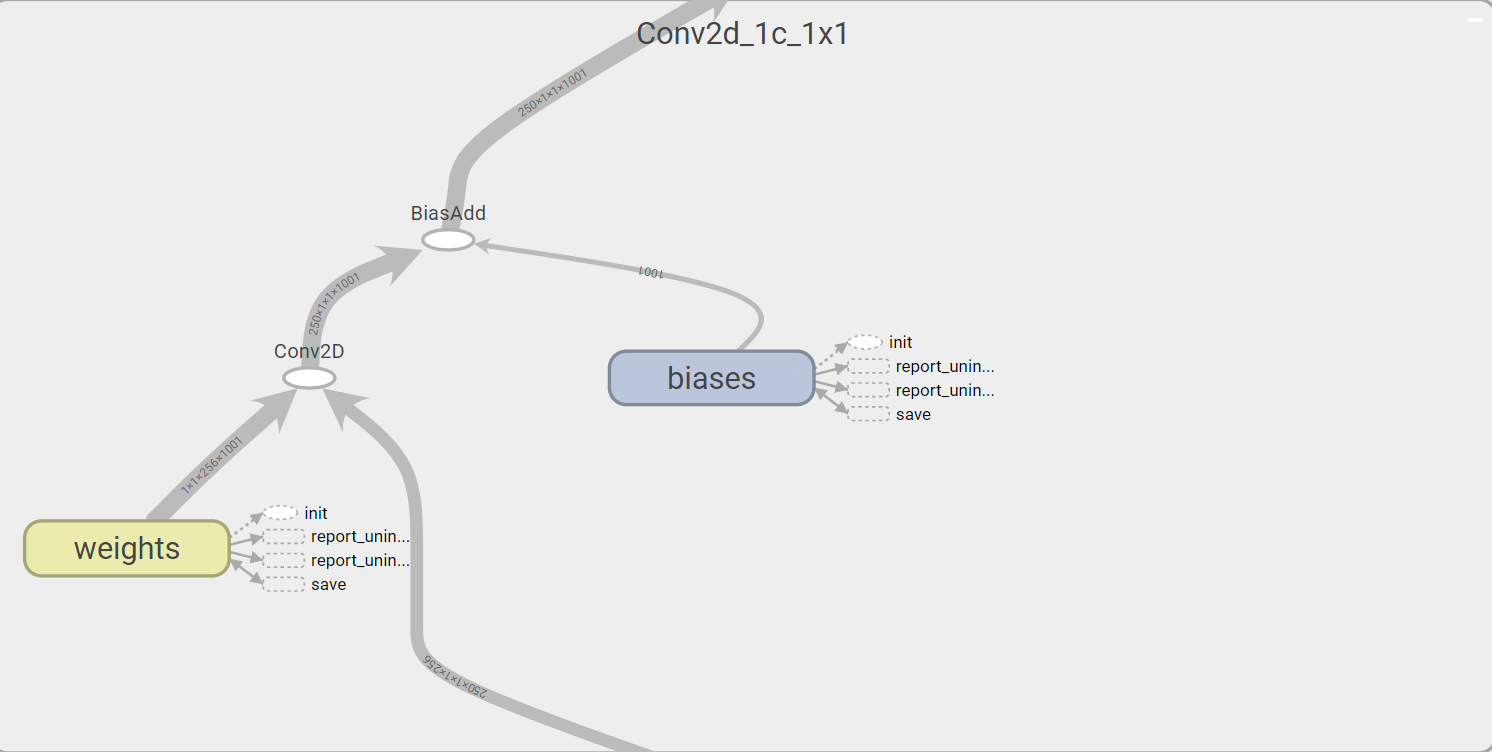}
\caption{Floating Point}
\end{subfigure}
\hfill
\begin{subfigure}[h]{0.8\linewidth}
\includegraphics[width=\linewidth]{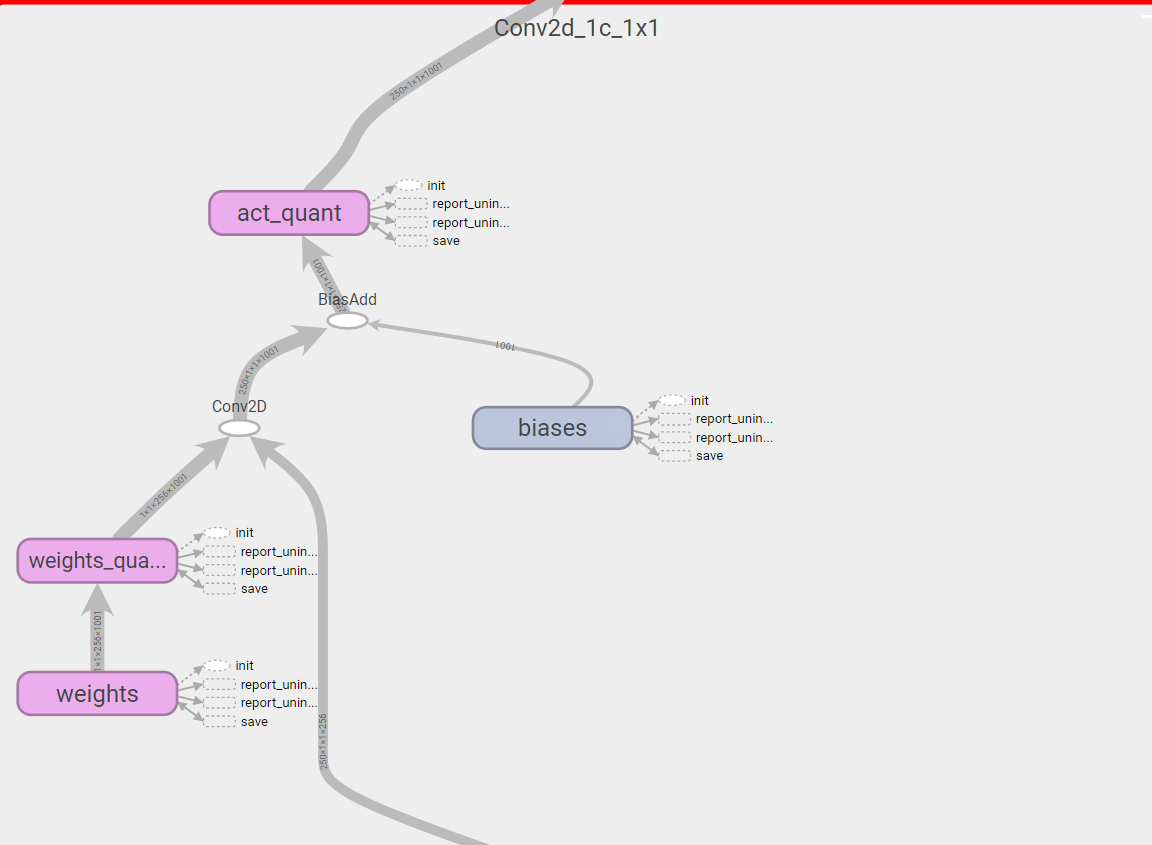}
\caption{Fixed Point}
\end{subfigure}%
\caption{Convolutional layer: Before and After Graph Transformation}
\label{fig:ConvLayerTransform}
\end{figure}

 The steps involved in training a quantized model are:
 \begin{enumerate}
 \item (Recommended): Fine tune from a floating point saved model: Start with a floating point pre-trained model or alternately train from scratch
 \item Modify Estimator to add quantization operations: Add fake quantization operations to the model using the quantization rewriter at tf.contrib.quantize
 \item Train model: At the end of this process, we have a savedmodel  with quantization information (scale, zero-point) for all the quantities of interest. (weights and activations) 
 
 \item Convert model: The savedmodel with range information is transformed into a flatbuffer file using the tensorflow converter (TOCO) at: tf.contrib.lite.toco\_convert. This step creates a flatbuffer file that converts the weights into integers and also contains information for quantized arithmetic with activations
 \item Execute model: The converted model with integer weights can now be executed using the TFLite interpreter which can optionally execute the model in custom accelerators using the NN-API. One can also run the model on the CPU.
\end{enumerate} 
 A simple example showing the graph transformation for a convolutional layer is shown in figure \ref{fig:ConvLayerTransform}.
 
 \subsubsection{Operation Transformations for Quantization}
\begin{figure*}
[!htbp]
\begin{center}
    \includegraphics[width=0.8\linewidth]{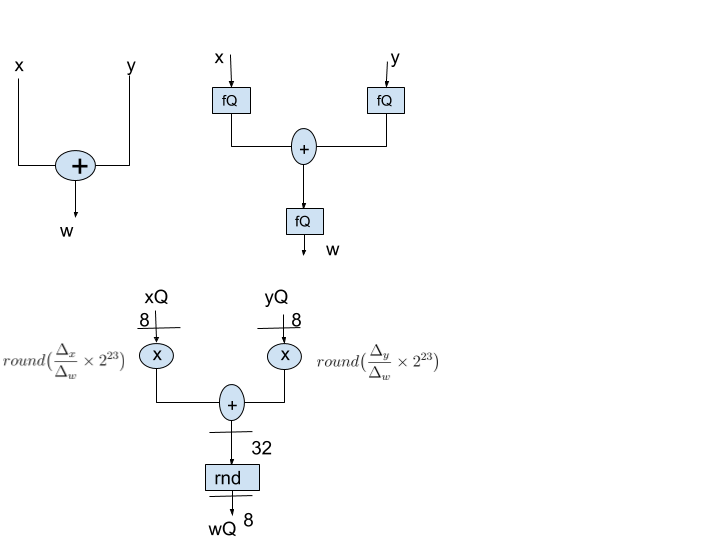}
\end{center}
   \caption{Fixed point transformation of element-wise add}
\label{fig:RescaleAdd}
\end{figure*}

\begin{figure*}
[!htbp]
\begin{center}
    \includegraphics[width=0.8\linewidth]{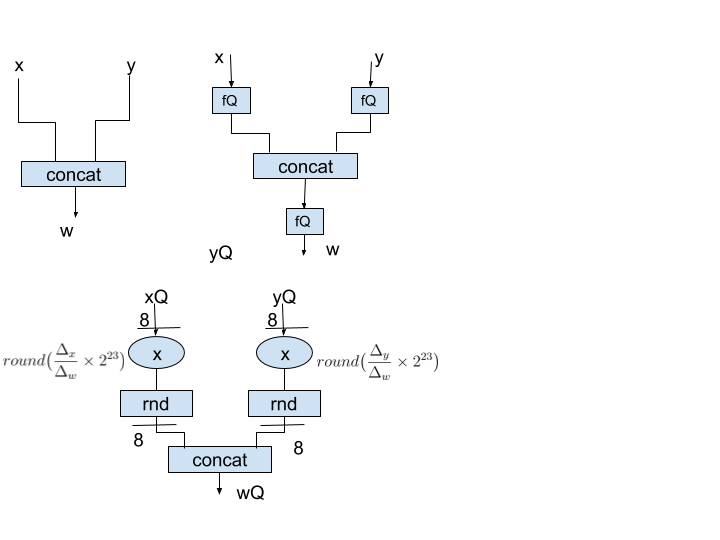}
\end{center}
   \caption{Fixed point transformation of concat}
\label{fig:RescaleConcat}
\end{figure*}

 It is important to ensure that all quantization related artifacts are faithfully modeled at training time. This can make trivial operations like addition, figure \ref{fig:RescaleAdd} and concatenation , figure \ref{fig:RescaleConcat}  non-trivial due to the need to rescale the fixed point values so that addition/concatenation can occur correctly. 
 
In addition, it is important to ensure that fusion of operations at inference time is modeled correctly during training. For example, consider an add followed by a ReLU operation. In this case, one can fuse the addition and the ReLU operation at inference time in most platforms. To match this, fake quantization operations should not be placed between the addition and the ReLU operations.

\subsubsection{Batch Normalization}
In this section, we describe several strategies for quantizing batch normalization layers. In section \ref{TrnPractices} and show that batch normalization with correction and freezing provides the best accuracy.

Batch normalization \cite{IoffeS15}, is a popular technique that normalizes the activation statistics at the output of every layer, reducing dependencies across layers while significantly improving model accuracy. 

Batch normalization is defined by the following equations:
\begin{equation}
  x_{bn}=\gamma\left(\frac{x -\mu_B}{\sigma_B}\right)+\beta
\end{equation}
for training and
\begin{equation}
  x_{bn}=\gamma\left(\frac{x - \mu}{\sigma}\right)+\beta
\end{equation}
for inference.

Where $ \mu_B  $ and $ \sigma_B $ are the batch mean and standard deviations. $\mu $ and  $ \sigma $ are the long term mean and standard deviations and
are computed as moving averages of batch statistic during training. 

For inference, we fold the batch normalization into the weights as defined by equations \ref{eq:1} and \ref{eq:2}. Therefore, at inference there is no explicit batch normalization.
The weights and biases are modified to account for batch normalization instead:
\begin{align} 
W_{inf}&= \frac{\gamma W}{\sigma} \label{eq:1} \\ 
Bias_{inf} &= \beta - \frac{\gamma \mu}{\sigma} \label{eq:2}
\end{align}

\begin{figure}
[!htbp]
\begin{center}
    \includegraphics[width=1.1\linewidth]{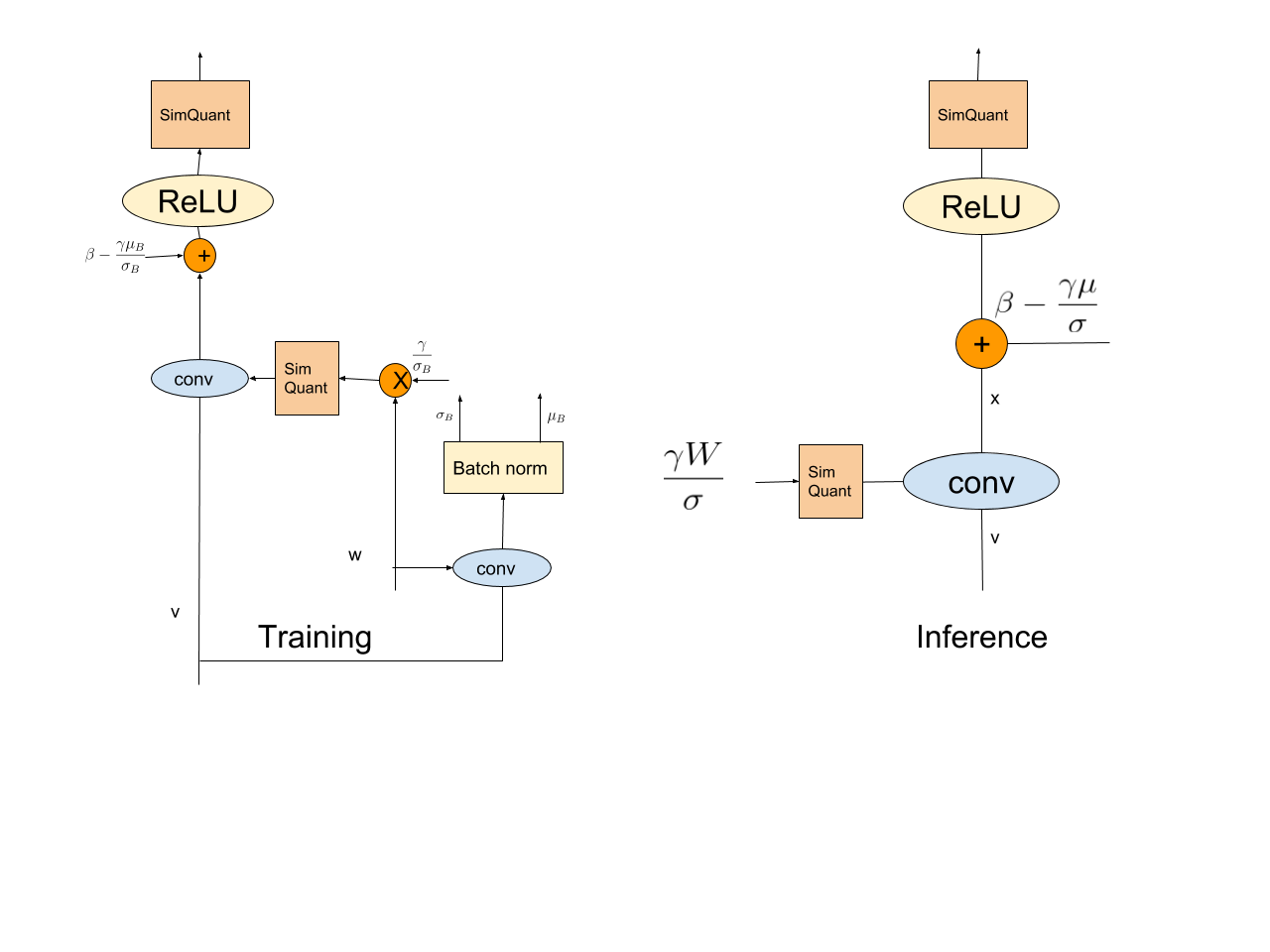}
\end{center}
   \caption{Baseline approach for folding batch norms for quantized inference}
\label{fig:NaiveBN}
\end{figure}

For quantized inference, we first consider folding batch norms and training them as shown in figure \ref{fig:NaiveBN}. We note that batch normalization uses batch statistics during training, but uses long term statistics during inference. Since the batch statistics vary every batch,
this introduces undesired jitter in the quantized weights and degrades the accuracy of quantized models. (green curve in \ref{fig:MobV1BNfreeze} and \ref{fig:EMACorrBN})
A simple solution would be to switch to using long term moving averages during training, however, this eliminates batch normalization (i.e the mean and variance used do not correspond to the batch statistics) and causes instability in training. The graph rewriter implements a solution that eliminates the mismatch between training and inference with batch normalization (see figure \ref{fig:CorrBN}):

\begin{figure*}
[!htbp]
\begin{center}
    \includegraphics[width=1.3\linewidth]{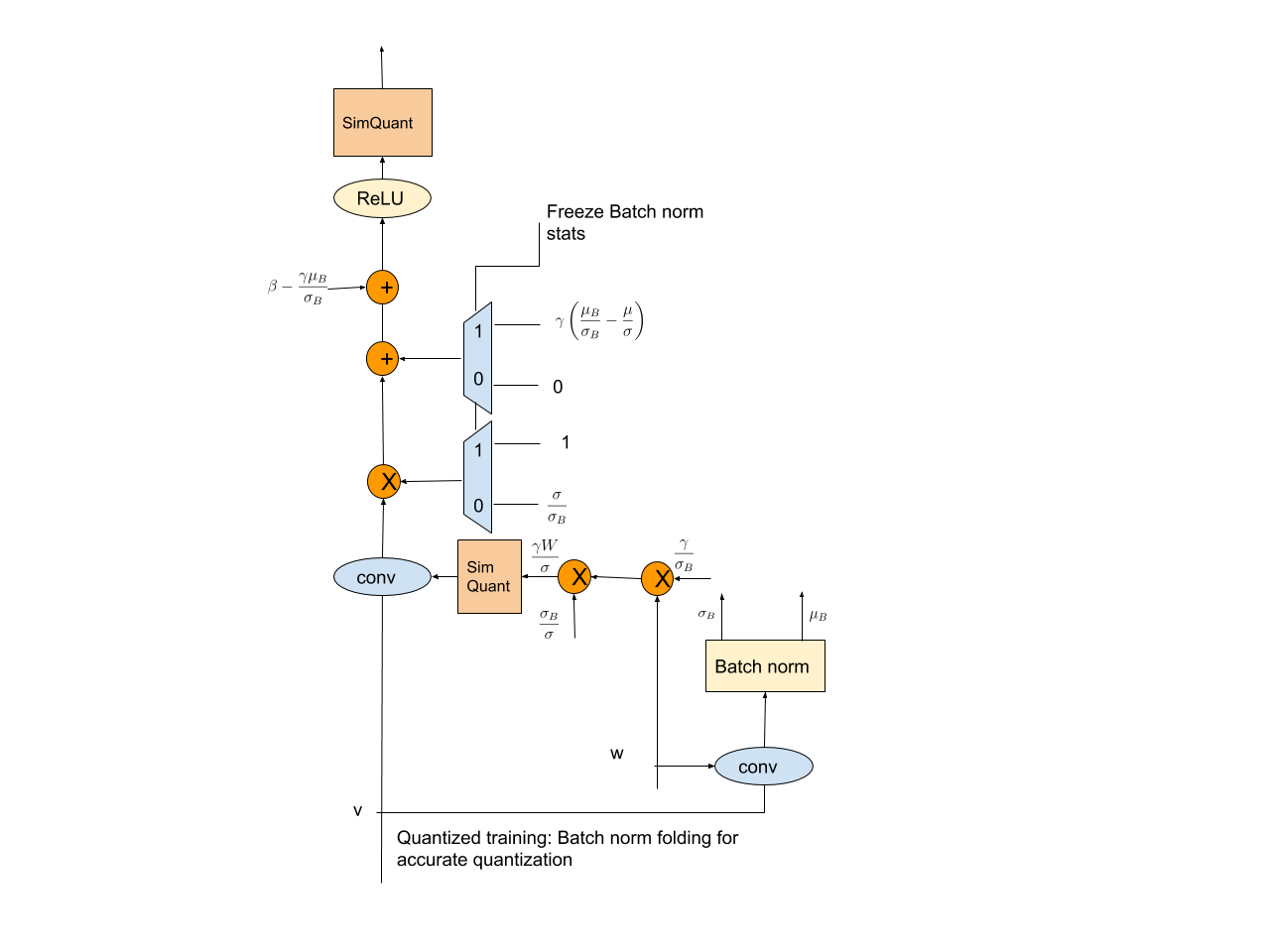}
\end{center}
   \caption{Folding Batch normalization layers for improved performance}
\label{fig:CorrBN}
\end{figure*}

\begin{enumerate}
\item We always scale the weights with a correction factor to the long term statistics prior to quantization. This ensures that there is no jitter in the quantized weights due to batch to batch variation.
\begin{align}
   c&=\frac{\sigma_B}{\sigma} \\
    w_{corrected}&=c \times \frac{\gamma W}{\sigma_B}
\end{align}

\item During the initial phase of training, we undo the scaling of the weights so that outputs are identical to regular batch normalization. We also modify the bias terms correspondingly.
\begin{align}
  y&=conv(Q(w_{corrected}),x) \\ 
  y_{corrected}&=y/c \\
  bias &= \beta -\gamma\mu_B/\sigma_B \\
  bias_{corrected}&= 0
\end{align}
\item After sufficient training, switch from using batch statistics to long term moving averages for batch normalization, using the optional parameter freeze\_bn\_delay  in \verb create_experimental_training_graph()   (about 300000 steps in figure \ref{fig:EMACorrBN} and 200000 in figure \ref{fig:MobV1BNfreeze}). Note that the long term averages are frozen to avoid instability in training.
This corresponds to the normalization parameters used at inference and provides stable performance.
\begin{align}
  y&=conv(Q(w_{corrected}),x) \\ 
  y_{corrected}&=y \\
  bias &= \beta -\gamma\mu_B/\sigma_B \\
  bias_{correction}&= \gamma(\mu_B/\sigma_B-\mu/\sigma) 
\end{align}

\end{enumerate}

\subsubsection{Experiments}
\begin{figure}
[!htbp]
\begin{center}
    \includegraphics[width=1.0\linewidth]{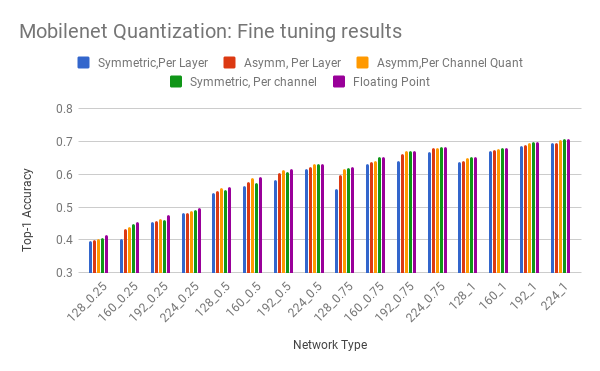}
\end{center}
   \caption{Comparison of quantization-aware training schemes:Mobilenet-v1}
\label{fig:QtzTrnMobilenetV1}
\end{figure}

\begin{figure}
[!htbp]
\begin{center}
    \includegraphics[width=12cm,height=6cm]{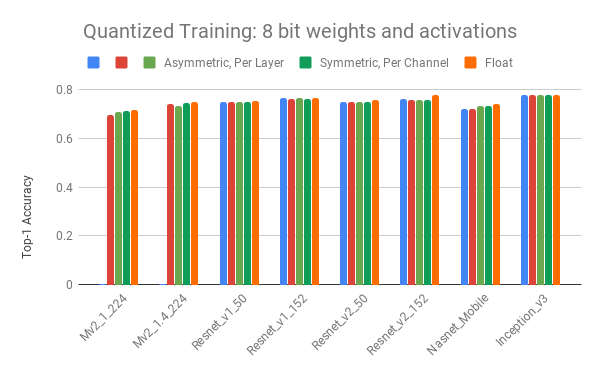}
\end{center}
   \caption{Comparison of quantization-aware training schemes}
\label{fig:QtzTrnSurvey}
\end{figure}

Quantization aware training closes the gap to floating point accuracy, even for per-layer quantization of weights. We repeat the same experiments for quantized weights and activations with training, starting from a floating point check-point and with batch normalization freezing and obtain the results shown in figures \ref{fig:QtzTrnSurvey} and \ref{fig:QtzTrnMobilenetV1} and Table \ref{Table:QtzAwareTrn8}.

All the experiments have the following settings:
\begin{itemize}
    \item Fine tune from a floating point checkpoint, we used the models in \cite{SlimModel17}.
    \item Use Stochastic Gradient Descent for fine tuning, with a step size of 1e-5.
    
\end{itemize}

\begin{table}[!htbp]
\begin{center}
  \begin{tabular}{|l|p{2cm}|p{2cm}|p{2cm}|p{2cm}|p{2cm}|}
  \hline
  
Network & Asymmetric, per-layer (Post Training Quantization) & Symmetric , per-channel (Post Training Quantization) &
Asymmetric, per-layer (Quantization Aware Training) & Symmetric, per-channel (Quantization Aware Training)   & Floating Point \\ \hline
Mobilenet-v1\_1\_224 &0.001 &0.591 &0.70&0.707& 0.709 \\ \hline
Mobilenet-v2\_1\_224 &0.001 &0.698 &0.709&0.711& 0.719 \\ \hline
Nasnet-Mobile& 0.722&0.721 & 0.73 &0.73&0.74 \\ \hline
Mobilenet-v2\_1.4\_224 &0.004 &0.74 &0.735 &0.745& 0.749 \\ \hline
Inception-v3 & 0.78& 0.78 & 0.78 & 0.78&0.78 \\ \hline
Resnet-v1\_50 & 0.75 & 0.751 &0.75 &0.75& 0.752 \\ \hline
Resnet-v2\_50 & 0.75 & 0.75 &0.75 & 0.75&0.756 \\ \hline
Resnet-v1\_152 & 0.766 & 0.762 &0.765 &0.762& 0.768 \\ \hline
Resnet-v2\_152 & 0.761 & 0.76 &0.76 &0.76& 0.778 \\ \hline

 \end{tabular}
\end{center}
\caption{Quantization aware training provides the best accuracy and allows for simpler quantization schemes}
\label{Table:QtzAwareTrn8}    
\end{table}
We note that:
\begin{enumerate}
    \item Training closes the gap between symmetric and asymmetric quantization. 
    \item Training allows for simpler quantization schemes to provide close to floating point accuracy. Even per-layer quantization shows close to floating point accuracy (see column 4 in Table \ref{Table:QtzAwareTrn8})

\end{enumerate}

\subsubsection{Lower Precision Networks}
\label{sec:LowPrecNet}
We note that at 8-bits of precision, post training quantization schemes provide close to floating point accuracy. In order to better understand the benefits of quantization aware training, we perform experiments to assess performance at 4 bit quantization for weights and activations.

We perform the following experiments:
\begin{itemize}
\item \textbf{Experiment 1: Per-channel quantization is significantly better than per-layer quantization at 4 bits.}  We show that per-channel quantization provides big gains over per-layer quantization for all networks. At 8-bits, the gains were not significant as there were sufficient levels to represent the weights with high fidelity. At four bits, the benefits of per-channel quantization are apparent, even for post training quantization (columns 2 and 3 of Table \ref{Table:WeightQuantPerLayerPerChannel}).
\begin{table}[!htbp]
\begin{center}
  \begin{tabular}{|l|p{2cm}|p{2cm}|p{2cm}|p{2cm}|}
  \hline
\diagbox{Network}{QuantizationType}	& Asymmetric, per-layer (Post Training Quantization) & Symmetric,per-channel (Post Training Quantization) & Symmetric, per-channel (Quantization Aware Training) &Floating Point \\ \hline
Mobilenet\_v1\_1\_224 &		0.02& 0.001& 0.65&0.709\\ \hline
Mobilenet\_v2\_1\_224	&	0.001& 0.001 & 0.62&0.719\\ \hline
Nasnet\_Mobile&		0.001& 0.36&0.7&0.74 \\ \hline
Mobilenet\_v2\_1.4\_224&		0.001& 0.001& 0.704& 0.749\\ \hline
Inception-v3 & 0.5& 0.71 & 0.76 & 0.78 \\ \hline
Resnet\_v1\_50&		0.002& 0.54 &0.732 & 0.752\\ \hline
Resnet\_v1\_152&		0.001& 0.64&0.725&0.768\\ \hline
Resnet\_v2\_50&		0.18& 0.72&0.73&0.756 \\ \hline
Resnet\_v2\_152&		0.18& 0.74&0.74&0.778 \\ \hline

\end{tabular}
\end{center}
\caption{4 bit Weight Quantization: per-channel quantization outperforms per-layer quantization, with fine tuning providing big improvements.}
\label{Table:WeightQuantPerLayerPerChannel}    

\end{table}
\item \textbf{Experiment 2: Fine tuning can provide substantial accuracy improvements at lower bitwidths.} It is interesting to see that for most networks, one can obtain accuracies within 5\% of 8-bit quantization with fine tuning 4 bit weights (column 4 of Table \ref{Table:WeightQuantPerLayerPerChannel}). The improvements due to fine tuning are also more apparent at 4 bits. Note that activations are quantized to 8-bits in these experiments.

\item \textbf{Experiment 3: Lower precision activations:} We investigate the accuracies obtained with 4-bit activations for all layers with and without fine tuning. Note that activations are quantized on a per-layer basis. The weights are quantized at 8-bits of precision with per-channel granularity. We note that fine tuning improves accuracy in this case also. The losses due to activation quantization are more severe than that of weight quantization (see Table \ref{Table:ActQuant}). Note that the quantization granularity is different for activations and weights, so this is not a fair comparison of the impact of quantization. We hypothesize that quantizing activations introduces random errors as the activation patterns vary from image to image, while weight quantization is deterministic. This allows for the network to learn weight values to better compensate for the deterministic distortion introduced by weight quantization. 
 
\begin{table}[h]
\begin{center}
  \begin{tabular}{|l|p{2cm}|p{2cm}|p{2cm}|p{2cm}|}
  \hline
\diagbox{Network}{QuantizationType} & Post Training Quantization (8,4) &  Quantization Aware Training (8,4) & Quantization Aware Training (4,8)&  Floating Point \\ \hline
Mobilenet\_v1\_1\_224 &0.48&	0.64 &0.65& 0.709\\ \hline
Mobilenet\_v2\_1\_224	& 0.07&0.58 & 0.62&0.719\\ \hline
Resnet\_v1\_50&	 0.36& 0.58 &0.732&0.752\\ \hline
Nasnet\_Mobile& 0.04&0.4&0.7&0.74 \\ \hline
Inception\_v3 & 0.59 &0.74&0.76& 0.78 \\ \hline
\end{tabular}
\end{center}
\caption{4 bit Activation Quantization with and without fine tuning. Note that weights are quantized with symmetric per-channel quantization at 8-bits. We also show results for 4-bit per-channel quantization of weights with 8-bit activations to compare with 8-bit weights and 4-bit activations}
\label{Table:ActQuant}    

\end{table}

\end{itemize}

\section{Training best practices} \label{TrnPractices}
We experiment with several configurations for training quantized models:
Our first experiment compares stochastic quantization with deterministic quantization. Subsequently, we study if training a quantized model from scratch provides higher accuracies than fine tuning from a floating point model. We also evaluate different methods for quantizing batch normalization layers and show that batch normalization with corrections provides the best accuracy. We also compare schemes that average weights during training with no averaging. 
\begin{enumerate}
    
\item \textbf{Stochastic Quantization does not improve accuracy:} Stochastic quantization determines floating point weights that provide robust performance under stochastic quantization, which causes the quantized weights to vary from mini-batch to mini-batch. At inference, quantization is deterministic, causing a mismatch with training. We observe that due to this mis-match, stochastic quantization under-performs determinstic quantization (figure \ref{fig:StochCompare}), which can be compensated better during training.

\begin{figure}
[!h]
\begin{center}
    \includegraphics[width=1.0\linewidth]{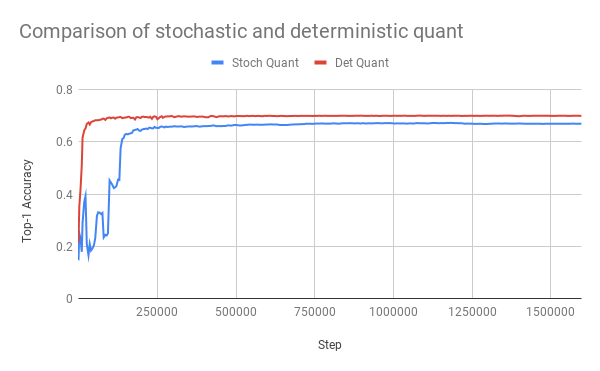}
\end{center}
   \caption{Comparison of stochastic quantization vs deterministic quantization during training}
\label{fig:StochCompare}
\end{figure}

\item \textbf{Quantizing a model from a floating point checkpoint provides better accuracy:} The question arises as to whether it is better to train a quantized model from scratch or from a floating point model. In agreement with other work \cite{Mishra17}, we notice better accuracy when we fine tune a floating point model as shown in figure \ref{fig:FineTunevsScratch}. This is consistent with the general observation that it is better to train a model with more degrees of freedom and then use that as a teacher to produce a smaller model (\cite{Hinton15}).

\begin{figure}
[!htbp]
\begin{center}
    \includegraphics[width=1.0\linewidth]{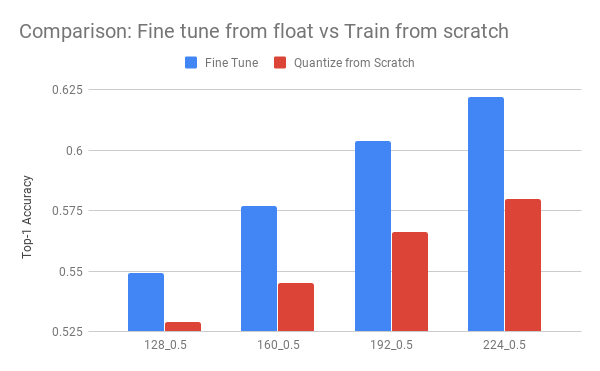}
\end{center}
   \caption{Fine tuning a floating point checkpoint provides better fixed point accuracy}
\label{fig:FineTunevsScratch}
\end{figure}

\item \textbf{Matching Batch normalization with inference reduces jitter and improves accuracy.} We show results for  two networks. In the first experiment (see figure  \ref{fig:MobV1BNfreeze}), we compare training with naive batch norm folding, batch renormalization and batch normalization with correction and freezing for Mobilenet-v1\_1\_224. We note stable eval accuracy and higher accuracy with our proposed approach. In the second experiment, we compare naive batch norm folding and batch normalization with correction and freezing for Mobilenet\_v2\_1\_224. We note that corrections stabilize training and freezing batch norms provides additional accuracy gain, seen after step 400000 in figure \ref{fig:EMACorrBN}.
\begin{figure*}
[!htbp]
\begin{center}
    \includegraphics[width=1.0\linewidth]{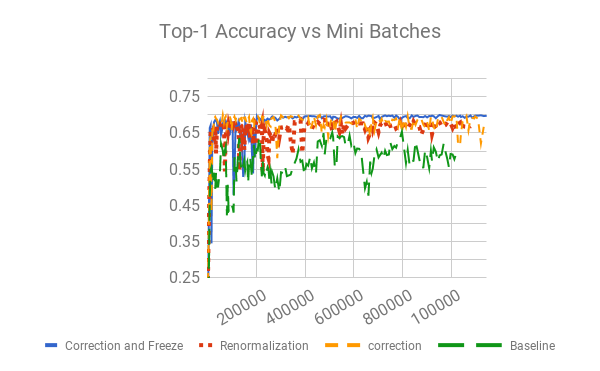}
\end{center}
   \caption{Mobilenet\_v1\_1\_224: Comparison of Batch normalization quantization schemes: Batch normalization without corrections (green) shows a lot of jitter due to the changing scaling of weights from batch to batch. Batch renormalization (red) improves the jitter, but does not eliminate it. Quantizing the weights using moving average statistics reduces jitter, but does not eliminate it (orange). Freezing the moving mean and variance updates after step 200000 allows for quantized weights to adapt to the batch norm induced scaling and provides the best accuracy with minimal jitter (blue curve). }
\label{fig:MobV1BNfreeze}
\end{figure*}

\begin{figure*}
[!htbp]
\begin{center}
    \includegraphics[width=1.0\linewidth]{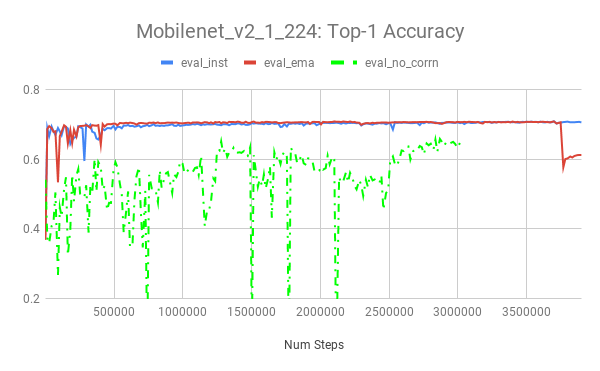}
\end{center}
   \caption{Mobilenet\_v2\_1\_224: Impact of batch normalization corrections and freezing on accuracy. Quantizing without corrections shows high jitter (green curve). Correction with freezing show good accuracy (blue and red curves).  The jitter in the eval accuracy drops significantly after moving averages are frozen (400000 steps). Note under performance of EMA weights (red curve) after sufficient training.}
\label{fig:EMACorrBN}
\end{figure*}

\item \textbf{Use Exponential moving averaging for quantization with caution.} Moving averages of weights \cite{Polyak} are commonly used in floating point training to provide improved accuracy \cite{EMATensorflow}.  Since we use quantized weights and activations during the back-propagation, the floating point weights converge to the quantization decision boundaries. Even minor variations in the floating point weights, between the instantaneous and moving averages can cause the quantized weights to be significantly different, hurting performance, see drop in accuracy for the EMA curve in figure \ref{fig:EMACorrBN}.

\end{enumerate}
\section{Model Architecture Recommendations}
In this section, we explore choices of activation functions and tradeoffs between precision and width of a network.

\begin{itemize}
\item \textbf{Do not constrain activation ranges:} One can get slightly better accuracy by replacing ReLU6 non-linearity with a ReLU and let the training determine the activation ranges (see  figure \ref{fig:ReLUvsReLU6})
\begin{figure}
[!htbp]
\begin{center}
    \includegraphics[width=1.0\linewidth]{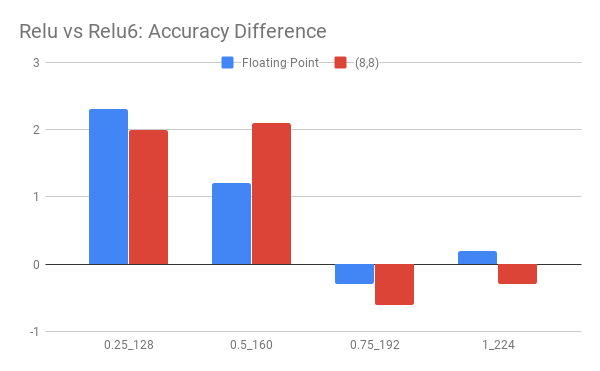}
\end{center}
   \caption{Accuracy improvement of training with ReLU over ReLU6 for floating point and quantized mobilenet-v1 networks.}
\label{fig:ReLUvsReLU6}
\end{figure}
\item \textbf{Explore tradeoff of width vs quantization:} An over-parameterized model is more amenable to quantization. Even for leaner architectures like mobilenet, one can tradeoff the depth multiplier with the precisions of the weights. We compare the accuracies obtained with 4 bit per-channel quantization of weights with 8-bit quantization across different depth multipliers in figure \ref{fig:WidthVsPrecision}. Note that this comparison allows us to evaluate a depth vs quantization tradeoff (see \cite{WRPN}). It is interesting to see that one can obtain a further 25\% reduction in the model size for almost the same accuracy by moving to 4 bit precision for the weights.
\begin{figure}
[!htbp]
\begin{center}
    \includegraphics[width=1.0\linewidth]{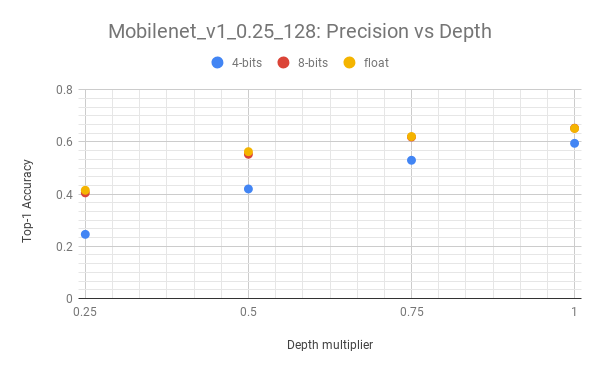}
\end{center}
   \caption{Width vs Precision tradeoff, illustrated for Mobilenet-v1\_0.25\_128, per-channel quantization of weights}
\label{fig:WidthVsPrecision}
\end{figure}

\end{itemize}
\section{Run-time measurements}
\label{sec:RTM}
We measure the run-times (Table \ref{Table:RunTimeResults}) on a single large core of the Google Pixel 2 device for both floating point and  quantized models. We also measure run-times using the Android NN-API on Qualcomm's DSPs. We see a speedup of 2x to 3x for quantized inference compared to float, with almost 10x speedup with Qualcomm DSPs.
\begin{table}[!htbp]
\begin{center}
  \begin{tabular}{|l|p{2cm}|p{2cm}|p{2cm}|}
  \hline
  \diagbox{Network}{Inference Platform} & Floating point(CPU) & Fixed point (CPU) & Fixed point (HVX, NN-API) \\ \hline
    Mobilenet\_v1\_1\_224 & 155 & 68 & 16 \\ \hline
    Mobilenet\_v2\_1\_224& 105& 63 & 15.5  \\ \hline
    Mobilenet\_v1\_1\_224\_SSD & 312 & 152 & \\ \hline
    Inception\_v3 & 1391 & 536  &   \\ \hline
    Resnet\_v1\_50  & 874 & 440 &  \\ \hline
    Resnet\_v2\_50  & 1667 & 1145 &  \\ \hline  
    Resnet\_v1\_152  & 2581 & 1274 &  \\ \hline  
    Resnet\_v2\_152  & 4885 & 3240 &  \\ \hline  

  \end{tabular}
\end{center}
\caption{Inference time measurements on Pixel2 phone in milliseconds on a single large core.}
\label{Table:RunTimeResults}    
\end{table}

\newpage

\section{Neural network accelerator recommendations}
\label{sec:NNRec}
In order to fully realize the gains of quantized networks, we recommend that neural network accelerators consider the following enhancements:
\begin{enumerate}
\item \textbf{Aggressive operator fusion:} Performing as many operations as possible in a single pass can lower the cost of memory accesses and provide significant improvements in run-time and power consumption.
\item \textbf{Compressed memory access:} One can optimize memory bandwidth by supporting on the fly de-compression of weights (and activations). A simple way to do that is to support lower precision storage of weights and possibly activations.
\item \textbf{Lower precision arithmetic:} One can obtain further acceleration by supporting a range of precisions for arithmetic. Our recommendation is to support 4,8 and 16-bit weights and activations. While 4 and 8-bit precisions are sufficient for classification, higher precision support is likely needed for regression applications, like super-resolution and HDR image processing.
\item \textbf{Per-layer selection of bitwidths:} We expect that many layers of a network can be processed at lower precision. Having this flexibility can further reduce model size and processing time.
\item \textbf{Per-channel quantization:} Support for per-channel quantization of weights is critical to allow for: \begin{enumerate}\item  Easier deployment of models in hardware, requiring no hardware specific fine tuning.
\item Lower precision computation.
\end{enumerate}
\end{enumerate}

\section{Conclusions and further work}
Based on our experiments, we make the following conclusions:
\begin{itemize}
\item \textbf{Quantizing models}
\begin{enumerate}
\item Use symmetric-per-channel quantization of weights with post training quantization as a starting point. Optionally fine tune if there is an accuracy drop.

\item  Quantization aware training can narrow the gap to floating point accuracy and in our experiments, reduce the gap to within 5\% of 8-bit quantized weights, even when all layers are quantized to 4 bits of precision.

\end{enumerate}

\item \textbf{Performance}
\begin{enumerate}
\item Quantized inference at 8-bits can provide 2x-3x speed-up on a CPU and close to 10x speedup compared to floating point inference on specialized processors optimized for low precision wide vector arithmetic, like the Qualcomm DSP with HVX.
\item One can obtain a model size reduction of 4x with no accuracy loss with uniform quantization. Higher compression can be obtained with non-uniform quantization techniques like K-means (\cite{HanMD15}).
\end{enumerate}

\item \textbf{Training Techniques} 
\begin{enumerate}
    \item  Quantization aware training can substantially improve the accuracy of models by modeling quantized weights and activations during the training process.
    \item It is critical to match quantized inference with the forward pass of training.
\item Special handling of batch normalization is required to obtain improved accuracy with quantized models.
\item Stochastic quantization during training underperforms deterministic quantization. 
\item Exponential Moving Averages of weights may under-perform instantaneous estimates during quantization aware training and must be used with caution.
\end{enumerate}
\item \textbf{Model architectures for quantization}
\begin{enumerate}
\item There is a clear tradeoff between model size and compressibility. Larger models are more tolerant of quantization error.
\item Within a single architecture, one can tradeoff feature-maps and quantization, with more feature maps allowing for lower bitwidth kernels.

\item One can obtain improved accuracy by not constraining the ranges of the activations during training and then quantizing them, instead of restricting the range to a fixed value. In our experiments, it was better to use a ReLU than a ReLU6 for the activations.
\end{enumerate}
\end{itemize}

Going forward, we plan to enhance our automated quantization tool to enable better quantization of networks by investigating the following areas:
\begin{enumerate}

\item Regularization techniques to better control the dynamic ranges of weights and activations can provide further improvements.

\item Distilled training to further improve the accuracy of quantized models \cite{AMishra17}.

\item Per-layer quantization of weights and activations to provide further compression and performance gains on hardware accelerators. Reinforcement learning has been applied successfully towards this problem in \cite{AMCSongHan}.
    
\end{enumerate}

\section{Acknowledgements}
 This work builds on the quantization scheme first introduced in \cite{BJacob17}. Suharsh Sivakumar developed the \cite{TFQuantize} tool used for the quantization experiments, which extend the capabilities first described in \cite{BJacob17}. Rocky Rhodes  provided the performance measurement numbers for the models. We would also like to thank Cliff Young, Song Han, Rocky Rhodes and Skirmantas Kligys for their useful comments. Pete Warden provided useful input on the scope of the paper and suggested several experiments included in this document.

\bibliographystyle{ieeetr}
\bibliography{egbib}
\newpage
\appendix
\begin{appendix}
\section{Impact of Batch Normalization on Quantization}
To understand the impact of batch normalization on the dynamic range of the folded weights (W), we consider the following metrics:

\begin{enumerate}
\item \textbf{SQNR:} We calculate the Signal to quantization noise ratio defined as: $$ SQNR=10log_{10}\frac{\sum W^2 }{\sum(W-SimQuant(W))^2} $$ for different quantization schemes.
We note that per-channel quantization provides significant improvement in SQNR over per-layer quantization, even if only symmetric quantization is used in the per-channel case.

\begin{figure}[!htbp]
\begin{subfigure}[h]{0.5\linewidth}
\includegraphics[width=\linewidth]{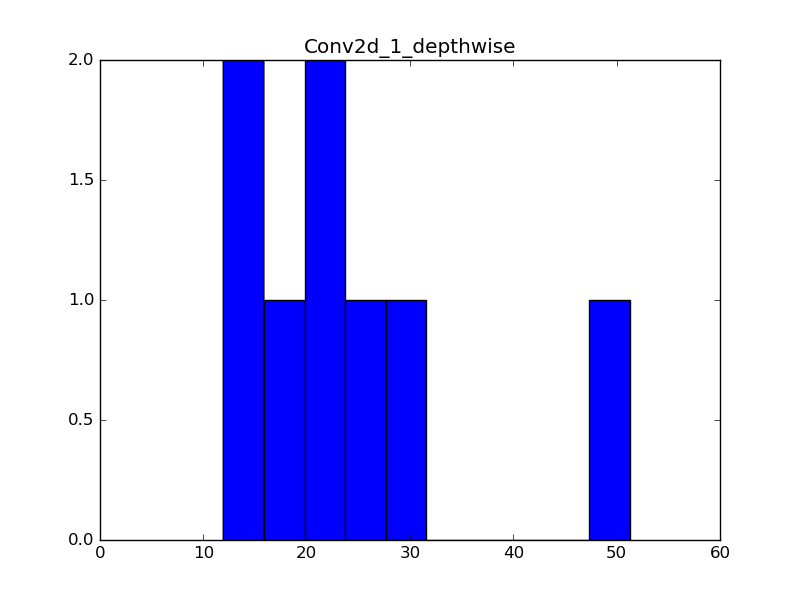}
\caption{Asymmetric, per-layer Quant}
\end{subfigure}
\hfill
\begin{subfigure}[h]{0.5\linewidth}
\includegraphics[width=\linewidth]{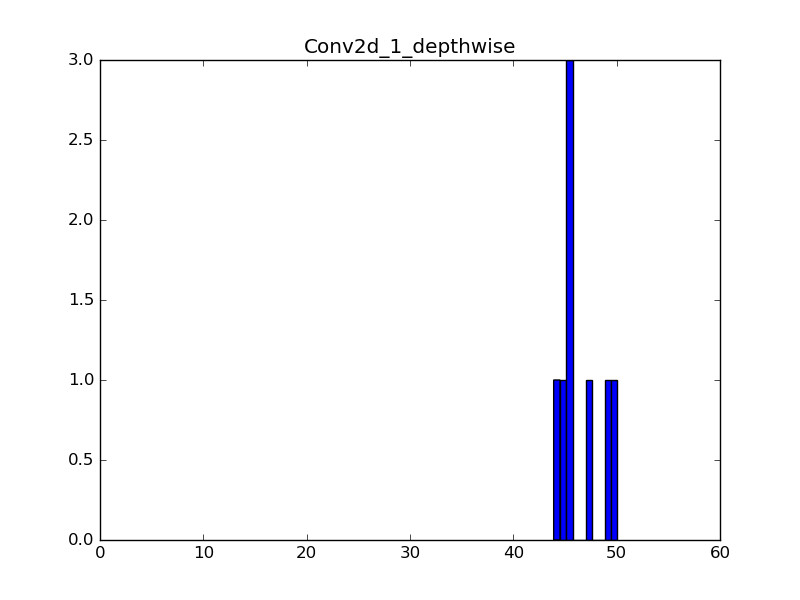}
\caption{Symmetric, per-channel Quant}
\end{subfigure}%
\caption{Histogram of the SQNR per output feature map (in dB on the x-axis), showing the number of kernels for each SQNR bin for different weight quantization schemes for layer:Conv2d\_1\_depthwise, Mobilenet\_v1\_0.25\_128. The total number of kernels is 8.}
\label{fig:Conv2d_1_depthwise_0.25_128_sqnr}
\end{figure}

\begin{figure}[!htbp]
\begin{subfigure}[h]{0.5\linewidth}
\includegraphics[width=\linewidth]{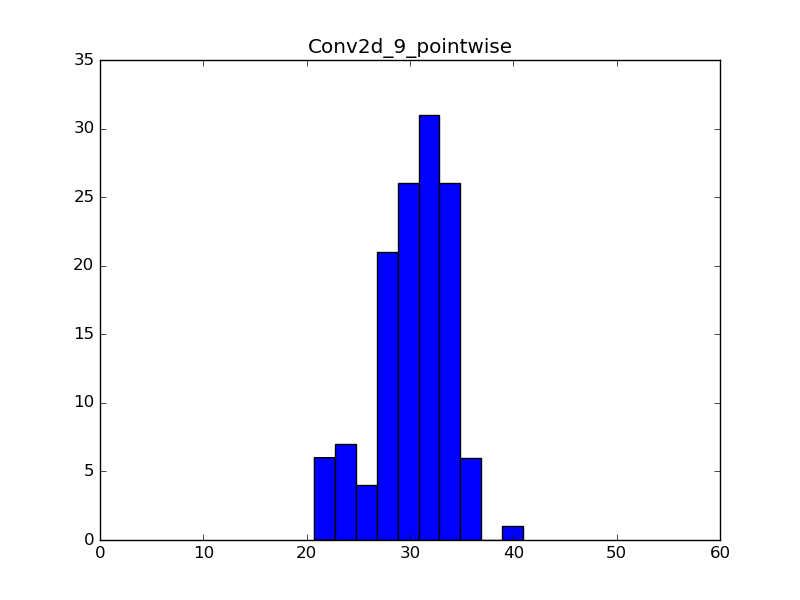}
\caption{Asymmetric, per-layer Quant}
\end{subfigure}
\hfill
\begin{subfigure}[h]{0.5\linewidth}
\includegraphics[width=\linewidth]{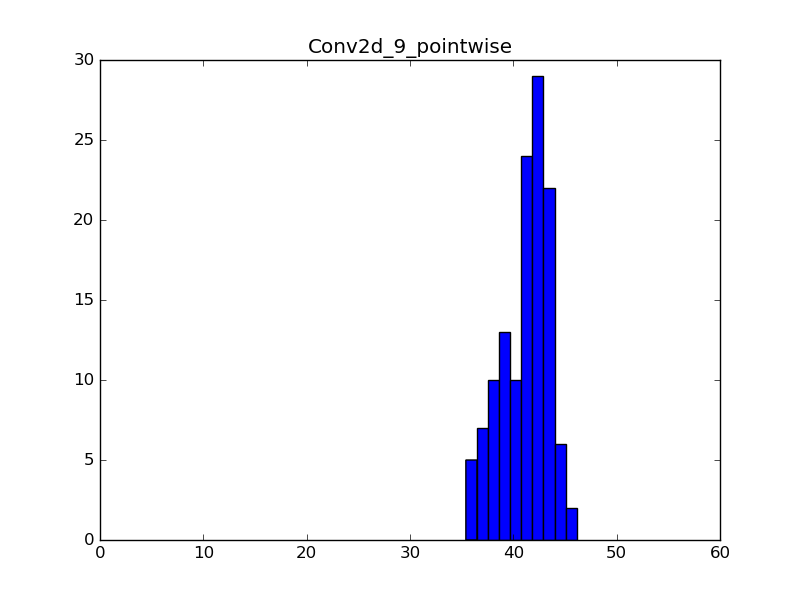}
\caption{Symmetric, per-channel Quant}
\end{subfigure}%
\caption{Histogram of the SQNR per output feature map (in dB on the x-axis), showing the number of kernels for each SQNR bin for different weight quantization schemes for layer:Conv2d\_9\_pointwise, Mobilenet\_v1\_0.25\_128. The total number of kernels is 128. }
\label{fig:Conv2d_9_pointwise_0.25_128_sqnr}
\end{figure}

\item \textbf{Weight Power Distribution:} We also plot the distribution of the sample weights, normalized by the average power, i.e we plot
 $$ histogram\big(\frac{W^2}{E[W^2]}\big) $$ for the weights before and after folding. We note that after folding, there are much larger outliers which severely degrade performance. 
\begin{figure}
[!htbp]
\begin{center}
    \includegraphics[width=0.8\linewidth]{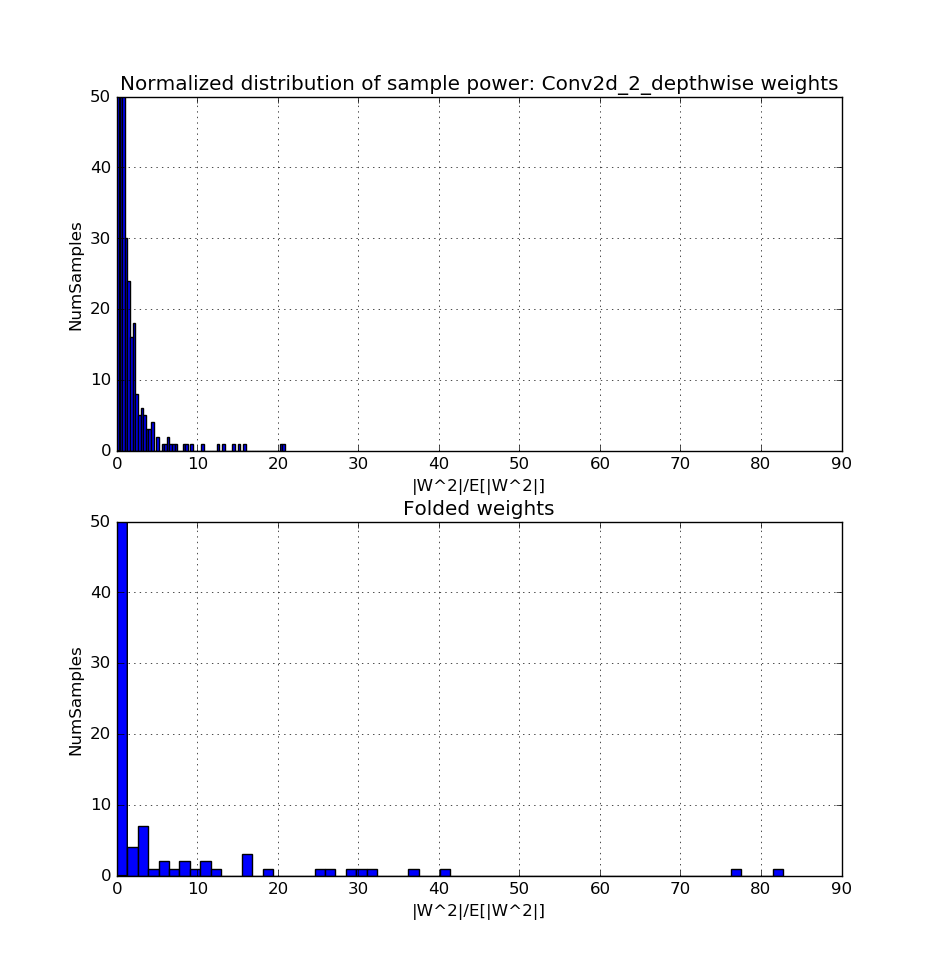}
\end{center}
   \caption{Weight histograms with and without folding:  Mobilenet\_V1\_1\_224, conv2d\_2\_depthwise, note the long tails of the distribution for folded weights.}
\label{fig:Conv2d_2_depthwise_224_hist}
\end{figure}

\end{enumerate}

\end{appendix}

\end{document}